\documentclass[10pt, a4paper]{article}

\usepackage{lrec-coling2024} % this is the new style
\usepackage{gb4e}

\title{The Syntactic Acceptability Dataset (Preview):\ A Resource for Machine Learning and Linguistic Analysis of English}

\name{Tom S Juzek} 

\address{Florida State University \\
         Tallahassee, FL, USA \\
         tjuzek@fsu.edu \\}

\abstract{
We present a preview of the Syntactic Acceptability Dataset, a resource being designed for both syntax and computational linguistics research. In its current form, the dataset comprises 1,000 English sequences from the syntactic discourse:\ Half from textbooks and half from the journal \textit{Linguistic Inquiry}, the latter to ensure a representation of the contemporary discourse. Each entry is labeled with its grammatical status (``well-formedness'' according to syntactic formalisms) extracted from the literature, as well as its acceptability status (``intuitive goodness'' as determined by native speakers) obtained through crowdsourcing, with highest experimental standards. Even in its preliminary form, this dataset stands as the largest of its kind that is publicly accessible. We also offer preliminary analyses addressing three debates in linguistics and computational linguistics:\ We observe that grammaticality and acceptability judgments converge in about 83\% of the cases and that ``in-betweenness'' occurs frequently. This corroborates existing research. We also find that while machine learning models struggle with predicting grammaticality, they perform considerably better in predicting acceptability. This is a novel finding. Future work will focus on expanding the dataset.
 \\ \newline  }

\begin{document}

\maketitleabstract

\section{Introduction}

One of the primary goals of syntactic theory is to identify the principles and processes that dictate the structure of sequences in a particular language and in human language in general. Syntax is primarily concerned with describing, explaining, and predicting the grammatical status of these sequences, particularly distinguishing between grammatical and ungrammatical sequences. Chomsky refers to these as members of the sets G and G', respectively \cite{chomsky1975logical}. Crucially, syntacticians focus on linguistic competence, which is a speaker's often implicit knowledge of a language \cite{chomsky1965aspects}.  

To build formalisms, syntacticians rely on various kinds of data, with a significant emphasis on their own expert judgments, often referred to as \textit{grammaticality judgments} \cite{schutze1996empirical, francis2021gradient}. These judgments are obtained when experts carefully examine and contrast linguistic sequences, determining whether a given sequence aligns with their grammatical formalisms. During this evaluation, linguists abstract away from extra-grammatical factors, such as memory limitations. This is illustrated in Sequence~\ref{ex:mce}, taken from \citet{chomsky1963introduction}. 

\begin{exe}
 \ex The rat the cat the dog chased killed ate the malt.
 \label{ex:mce}
\end{exe}

\noindent In recent years, sequences and their grammaticality evaluations have become increasingly accessible. The largest source of such data to date is the Corpus of Linguistic Acceptability (CoLA; \citealp{warstadt2019neural}), which contains more than 10,000 sequences and their respective grammatical statuses (see Section~\ref{sec:graacc} for why CoLA contains grammaticality judgments instead of acceptability judgments, as per standard usage in linguistics). The sequences in CoLA are sourced from syntax textbooks, and their grammaticality evaluations are provided by the authors of these textbooks (see the Appendix for examples).

\begin{table*}[!ht]
\begin{center}
\begin{tabularx}{420pt}{|c|c|c|c|c|X|}

      \hline
      \textbf{Gra} & \textbf{Acc} & \textbf{Nm-ac} & \textbf{Bi-ac} & \textbf{Src} & \textbf{Sequence} \\
      \hline
      0 & 1.35 & 0.06 & 0 & jl & John is too much to play with your kids old. \\
      ... & ... & ... & ... & ... & ... \\
      0 & 3.90 & 0.48 & 0 & tb & I assumed to be innocent. \\
      ... & ... & ... & ... & ... & ... \\
      1 & 6.89 & 0.99 & 1 & tb & I saw John on Sunday. \\
      \hline

\end{tabularx}
\caption{The structure of the our dataset, including labels for grammaticality, acceptability, normalized acceptability, binary acceptability, source (textbook or journal), and the sequence.}
\label{tab1}
 \end{center}
\end{table*}

\subsection{Issues surrounding Grammaticality}

Several issues arise when discussing grammaticality judgments. First, there is the matter of data adequacy and convergence. Sequence~\ref{ex:ungraacc}, taken from \citet{landau2007epp}, is labeled as ungrammatical by the original author from whom the sequence was sourced. However, most laypeople consider the sequence to be acceptable \cite[p.~207]{francis2021gradient}. 

\begin{exe}
 \ex *October 1st, he came back.  
 \label{ex:ungraacc}
\end{exe}

\noindent Furthermore, there is the question of \textit{gradience}. Is grammaticality a binary concept, or do degrees of (un)grammaticality exist \cite{chomsky1975logical,wasow2007gradient,francis2021gradient}? Considering a sequence such as Sequence~\ref{ex:gradience}, taken from \citet[p.~38]{francis2021gradient}, it becomes challenging to pinpoint which factors outside the traditional grammar influence the perception of the sequence as neither fully grammatical nor fully ungrammatical.

\begin{exe}
 \ex Olson brings to the table a great deal of experience. 
 \label{ex:gradience}
\end{exe} 

\noindent Thirdly, it has been observed that machine learning models struggle with the notion of grammaticality. \citet{warstadt2019neural} trained various LSTM models on CoLA and observed accuracy results well below 80\%. Given that contemporary models demonstrate high proficiency in various linguistic tasks \cite{devlin2018bert, tenney2019you}, the machine learning of grammaticality is of particular interest.

\section{Acceptability}\label{sec:graacc}

There are, however, many methods at the disposal of syntacticians to assist them in theory-building. These methods include self-paced reading tasks, EEG measurements, eye-tracking, and eliciting acceptability judgments. As to the latter, the linguistic literature defines acceptability judgments as non-expert, `naive' intuitions about the goodness of a sequence (see discussion in \citealp[pp.235-236]{haussler2020linguistic}, as well as various contributions in \citealp{schindler2020linguistic}, and see \citealp{etxeberria2018relating} and \citealp{schoenmakers2023linguistic} for further nuances). Acceptability judgments can be influenced by extra-grammatical factors \cite{schutze2020,lkeska2023panoramic}, which contrasts to grammaticality judgments, where experts abstract away from extra-grammatical factors as much as possible \cite{schutze1996empirical}. When carefully controlled and analyzed, acceptability judgments can serve as a proxy for grammaticality \cite{schutze2020,feldhausen2021revisiting}. To illustrate the distinction between the two concepts, reconsider the examples from the previous section. Sequence~\ref{ex:mce} is grammatical according to most syntactic frameworks, yet many native speakers find it unacceptable. Sequence~\ref{ex:ungraacc} is ungrammatical according to most frameworks, yet many native speakers find it acceptable. 

Acceptability data are reliable \cite{langsford2018quantifying} and instructive for linguistic purposes, and an increasing number of studies make use of them, where recent examples include \citet{fanselow2019cares}, \citet{hoot2021trace}, \citet{urtzi2022polarity}, and \citet{lami2022compound}. However, the collection of acceptability data is also resource-intensive. As a result, there are no large-scale datasets publicly available. To our knowledge, the largest datasets available are those by \citet{lau2017grammaticality} with 400 items, and \citet{warstadt2019neural} with 200 items. Others, for example \citet{sprouse2013comparison} for English, and \citet{chen2020assessing} for Mandarin, report on similar mid-sized datasets; these are, however, not publicly available. The primary objective of this study is to produce and offer a publicly available dataset on a large(r) scale. We present initial acceptability judgments for 1,000 items, with an eventual goal of scaling this to approximately 15,000 items. The dataset encompasses sequences, grammaticality judgments, acceptability judgments (raw, normalized, and converted), and encodes its source (textbook vs journal). Even this preliminary dataset addresses the three issues mentioned earlier:\ data convergence, gradience, and challenges in machine learning. We will delve deeper into these three topics in Section~\ref{sec:analyses}.

\section{Dataset Building}

Our data are taken from two sources, representing two conditions. The first condition, referred to as the `textbook condition', consists of 500 English sequences randomly sampled from CoLA, which itself is sourced from various syntax textbooks. The second condition, the `journal condition', comprises 500 English sequences randomly sampled from the data from \citet{juzek2020data}, who in turn sampled their items from the journal \textit{Linguistic Inquiry}. In terms of their structure, items from both conditions are similar: a sequence relevant to syntax is presented alongside a grammaticality judgment. However, a \textit{possible} difference lies in their grammatical status. It is anticipated that the sequences from textbooks are more foundational and well-established. Importantly, both sets of items are accompanied by grammaticality evaluations, as provided by their original sources. Examples of these sequences can be found in Table~\ref{tab1}. Examples of the data from which we sampled, that is examples from both CoLA and the data in \citet{juzek2020data}, can be found in the Appendix. 

Of the sequences from textbooks, 71.4\% were grammatical, compared to 67.4\% from the journal. Consequently, our dataset exhibits an imbalance. We opted to sample singletons rather than minimal pairs, primarily because many items in the literature are presented without counterparts. For detailed discussions on this choice, refer to discussions in \citet{warstadt2019neural} and \citet{juzek2020data}. 

\begin{figure}[!ht]
\begin{center}
\includegraphics[scale=0.3]{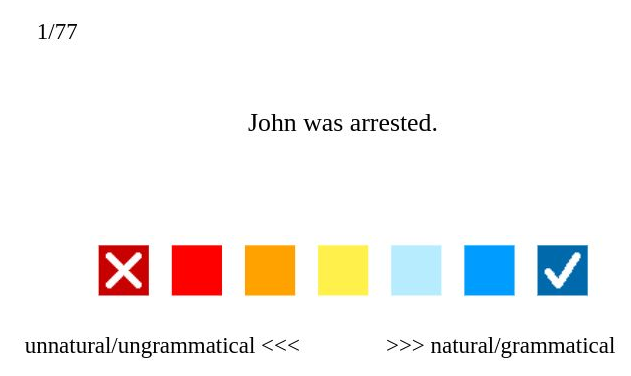} 
\caption{The interface of the judgment study.}
\label{ui}
\end{center}
\end{figure}

\begin{figure*}[!ht]
\begin{center}
\includegraphics[scale=0.5]{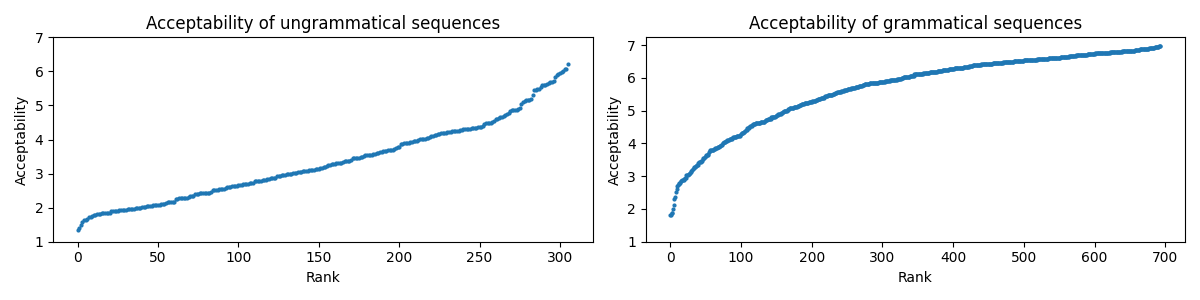} 
\caption{The items and their average acceptability ratings, sorted in ascending order, from unacceptable to acceptable. Left:\ Items evaluated as ungrammatical in the original source. Right:\ Items evaluated as grammatical.}
\label{plotgoodbad}
\end{center}
\end{figure*}

\subsection{Obtaining Acceptability Judgments}

Acceptability judgments were obtained through a self-hosted rating platform using the interface shown in Figure~\ref{ui}. Participants were crowdsourced via Prolific.com. A prerequisite for participation was that participants on Prolific had set their first language as American English. On average, participants were paid \$15/hr. After participants were provided with IRB information and instructions, they rated 77 items, 64 of which were critical items. We limited the number of items to avoid experimental fatigue. The first four items served as calibration items, taken from previous experiments to represent near-endpoints:\ two were unacceptable and two were acceptable. Participants then rated the remaining items. We opted not to include filler items since no distractor items were needed for our study, and the critical set covers all parts of the scale. 

For each participant, the 64 critical items were semi-randomly selected from the 1,000 items in our dataset, prioritizing items that had received the fewest ratings thus far. We interspersed four items testing language proficiency and five items testing general attention (of the sort "Please click on the leftmost button"). The platform also measured response times. If participants responded unrealistically fast, they received a warning. Those who were repeatedly non-cooperative were excluded immediately. After successfully completing the tasks, we collected basic demographics: gender, age, and first language. A total of 597 participants took part. We excluded users for the following reasons: less than 69 ratings were given (16 participants), unrealistically fast responses (2 participants), failing on language proficiency items (36 participants), failing on instructional items (2 participants), identifying as non-native speakers (2 participants). This commitment to quality is evident as the item with the lowest average rating had a score of 1.35 (an instructional item even averaged 1.03), while the highest-rated item had an average rating of 6.98. This indicates that participants utilized the entire rating scale. While under certain circumstances, fewer than ten items have been shown to give robust results \cite{mahowald2016snap}, we adhered to a more conservative N, with an average of over 30 ratings per item. In total, the dataset consists of 34,490 acceptability judgments, making it the most extensive publicly available dataset of its kind. 

\subsection{The Dataset}

The structure of the dataset is detailed in Table~\ref{tab1}. The dataset includes average acceptability ratings given on a 7-point scale. These ratings were then normalized to values between 0 and 1. For the purpose of binary classification, ratings between 0 and 0.5 were converted to 0, while ratings between 0.5 and 1 were converted to 1. In cases where ratings were exactly 0.5, we used the respective grammaticality value to determine the binary label. Figures~\ref{plotgoodbad} and \ref{plotgradience} illustrate the data distribution and structure, both of which will be discussed in the following section.

\section{Preliminary Analyses}\label{sec:analyses}

\subsection{Data Convergence}

83.3\% of all items share their grammaticality label and (to binary form converted) acceptability label. This rate is slightly higher in the textbook condition, at 85.8\%, and lower in the journal condition, at 80.8\%. These figures align with discussions in \citet{wasow2005intuitions} and \citet{gibson2013quantitative}, and with previous results in \citet{warstadt2019neural} and \citet{juzek2020data}. For an analysis of paired items, with a higher convergence rate, see \citet{sprouse2013comparison}. Moreover, the convergence rate for grammatical items (89.3\%) is considerably higher than for ungrammatical ones (69.6\%). This discrepancy is a novel finding and requires further investigation through a detailed item-by-item analysis:\ Apparently, there are numerous items that syntacticians label as ill-formed based on their formalisms, but which laypeople deem relatively acceptable. Sequence~\ref{ex:accbutung} serves as an example of this discrepancy. It was evaluated as ungrammatical in its original source, but received an average rating of 6.22 in our experiment. Moreover, the observed divergence rate of approximately 20\% underscores the idea that grammaticality and acceptability are indeed two distinct concepts. 

\begin{exe}
 \ex *John perfectly rolled the ball. \hspace{0.1cm} (6.22)
 \label{ex:accbutung}
\end{exe}

\subsection{Gradience}

\begin{figure}[!ht]
\begin{center}
\includegraphics[scale=0.5]{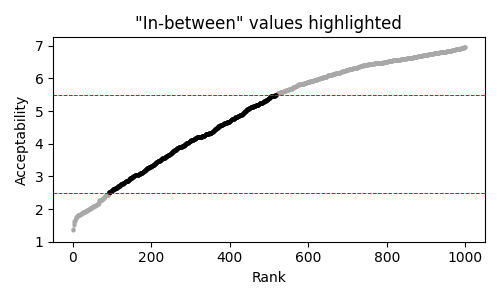} 
\caption{All items, with the mid-bin highlighted.}
\label{plotgradience}
\end{center}
\end{figure}

As consistently observed in the literature, acceptability exhibits a gradient nature (e.g.\ \citealp{featherston2005decathlon,wasow2007gradient, haussler2020linguistic}). The degree of this gradience is more pronounced than one might initially anticipate. In our results, when all items are ordered by rank in an ascending manner, as per Figure~\ref{plotgradience}, and when the initial few items with the lowest ratings are disregarded, there is a near-linear increase in acceptability. Interestingly, towards the higher end of the rating scale, the curve begins to resemble a saturation curve. This is in contrast to the sharper S-curve that one might expect. When the rating scale is divided into thirds, 28.3\% of all items are found in the middle bin. When the scale is divided into two bins:\ endpoint items (with ratings from 1 to 2.5 and 5.5 to 7) and in-between items (with ratings larger than 2.5 and smaller than 5.5), 42.6\% of all items are are found in the middle bin. This distribution is illustrated in Figure~\ref{plotgradience}. Our findings align with the discussion in \citet{francis2021gradient}. 

\subsection{Challenges in Machine Learning}\label{subsecml}

Transformers \cite{vaswani2017attention} demonstrate remarkable language abilities \cite{devlin2018bert}. \citet{hewitt2019structural} provided documentation of explicit syntactic representations. Conversely, however, \citet{chaves2020don}, \citet{chaves2021look}, and others, have delineated the challenges that deep learning encounters with syntactic constructions. Here, we ask how machine learning of grammaticality compares to machine learning of acceptability. Our dataset is relatively small for machine learning and can thus be viewed as a scarce-data learning scenario \cite{wang2020generalizing}. We fine-tuned Transformers on our data, with pre-trained models from \citet{wolf-etal-2020-transformers} (``bert-base-uncased''), for four conditions:\ 1) predicting grammaticality, 2) predicting acceptability, 3) predicting end-point acceptability (which excludes ``in-between'' items as per Figure~\ref{plotgradience}). Additionally, we include 4) a baseline condition where we sampled sentences from the Leipzig Corpora Collection for English \cite{goldhahn2012building} (labelled `good') and scrambled the word order of 500 sentences (labelled `bad'), then fine-tuned a Transformer to make predictions on these. Linear confusion matrices for these conditions are presented in Table~\ref{tab2}. 

As expected, the baseline model performs well, which may suggest that syntactic (un)acceptability cannot be solely predicted by word order. We observe that the models struggle with grammaticality, but they perform better on acceptability items. Furthermore, their performance on end-point acceptability is considerably better. These findings regarding acceptability are novel, but also limited in scope, and thus, warrant further, more in-depth research. Further, these findings also motivate the distinction between grammaticality and acceptability. 

\begin{table}[!ht]
\begin{center}
\begin{tabular}{|c|c|c|c|c|c|}

      \hline
      \textbf{Condition} & \textbf{tn} & \textbf{fp} & \textbf{fn} & \textbf{tp} & \textbf{Accu} \\
      \hline
      Grammatic. & 0 & 45 & 0 & 105 & 70\% \\
      \hline
      Acceptab. & 23 & 21 & 5 & 101 & 83\% \\
      \hline
      End-p.\ acc. & 9 & 3 & 2 & 73 & 94\% \\
      \hline
      Baseline & 39 & 0 & 4 & 107 & 97\% \\
      \hline

\end{tabular}
\caption{Linear confusion matrices for transformers, fine-tuned on our data in the different conditions, as per Section~\ref{subsecml}. Test data is 15\% of the dataset.}
\label{tab2}
 \end{center}
\end{table}

\section{Next Steps}

\subsection{Scaling}

While 1,000 items are a good start, the ideal situation for syntactic theory building would be that syntacticians can look up the acceptability of all relevant items. For this, we wish to expand our dataset to all items in \citet{warstadt2019neural} and \citet{juzek2020data}, resulting in a dataset of about 15,000 items. This would also help solidify our insights regarding the machine learning of acceptability. 

\subsection{Further annotations}

Further, ideally, syntacticians would not only be able to look up items but also search for syntactic constructions. This would require expert annotations for the items in the dataset. We are currently exploring possibilities to efficiently add such annotations. A scaled corpus with annotations for constructions could help with theory building and could inform research lines such as the work by \citet{bader2010toward} and \citet{bizzoni2020human}. 

Prosodic information could also be of interest for syntactic analysis. Recent research is underlining the relevance of prosody-syntax interactions, for work on such effects see e.g.\ \citet{wasow2015processing}, \citet{tang2021prosody}, \citet{gonzalez2022prosodic}, and \citet{gonzalez2024intonation}. Information-theoretic measures like surprisal and perplexity \cite{shannon1948mathematical} and advanced measures like lossy surprisal \cite{futrell2020lossy} could allow for advanced syntactic analyses. We also wish to add dependency parses \cite{tesniere2015elements,nivre2008algorithms} to further facilitate dependency-based syntactic research along the lines of \citet{gibson2000dependency}, \citet{liu2010dependency}. 

\subsection{Further analyses}

Thirdly, while we used response times for exclusions, we still need to do further analyses on the collected response times. For example, it could be interesting to see if there is a correlation between unacceptability and increased response times. Further, the setup allows for analyses such as the F\textsubscript{1}-score \cite{chinchor1992muc,van1979information} or the MCC \cite{chicco2020advantages}. An alternative line of analysis could involve a detailed examination of the `micro'-factors that underpin the observed `macro'-trends. This could be done by carefully analyzing a selection of items from the dataset. \citet{krielke2024cross} offers an exemplary analysis in this regard, which could serve as a model for our further investigations.

\section{Concluding Remarks}

We have introduced a preview of the Syntactic Acceptability Dataset, which comprises 1,000 sentences sourced from syntactic textbooks and the journal \textit{Linguistic Inquiry}. Each item in the dataset is accompanied by grammaticality evaluations and high-quality acceptability ratings. Even in its current form, this dataset is considerably larger than any other acceptability dataset currently available, and it has already provided insights into several debates. The dataset aids in understanding issues related to data convergence (with grammaticality and acceptability converging in about 83\% of cases, and a higher rate for textbook sequences), gradience (items with intermediate ratings are common), and machine learning challenges (grammaticality proves more difficult to predict than acceptability). In the next phase, we aim to expand the dataset, by adding more items, additional annotations, more advanced analyses, and further validate our preliminary findings.

\section*{Data Availability}

% import to tjuzek

The dataset and all relevant scripts are on Github: \href{https://www.github.com/tjuzek/sad}{github.com/tjuzek/sad}. 

%Output natbib command Old command
%(Eco, 1990) \citep \cite
%Eco, 1990 \citealp no equivalent
%Eco (1990) \citet \newcite
% (1990) \citeyearpar \shortcit

\section*{Acknowledgements}

The project received financial support from the Florida State University FYAP grant. Sincere appreciation is extended to the reviewers, as well as Elaine Francis and Tom Wasow for their valuable feedback.

\section*{Appendix}

In the following, we give an illustration of textbook items and journal items, to demonstrate that in structure, the sequences are similar. The items are used to analyze syntactic structures and the sequences come with an evaluation of their grammaticality (indicated by the asterisk or the lack thereof). 

Sequences~\ref{ex:textbookungr1} and \ref{ex:textbookungr2} are ungrammatical items taken from textbooks (from \citealp{kim2008english} and \citealp{baltin1991handbook}, respectively; we sampled them from CoLA), Sequences~\ref{ex:journalungr1} and \ref{ex:journalungr2} come from the journal \textit{Linguistic Inquiry} (both items come from \citealp{grosu2006reply}; we sampled them through \citealp{juzek2020data}). Similarly, Sequences~\ref{ex:textbookgr1} and \ref{ex:textbookgr2} are grammatical sequences from linguistics textbooks (from \citealp{adger2003core} and \citealp{sportiche2013introduction}, respectively) and Sequences~\ref{ex:journalgr1} and \ref{ex:journalgr2} are grammatical sequences from \textit{Linguistic Inquiry} (both items come from \citealp{basilico2003topic}). 

\begin{exe}
 \ex *You didn't leave, left you? 
 \label{ex:textbookungr1}
\end{exe}

\begin{exe}
 \ex *John seems will win. 
 \label{ex:textbookungr2}
\end{exe}

\begin{exe}
 \ex *John is too much to play with your kids old.
 \hspace{-0.4cm}
 \label{ex:journalungr1}
\end{exe}

\begin{exe}
 \ex *John is a more unusually than any of you is dressed student.
 \label{ex:journalungr2}
\end{exe}

\begin{exe}
 \ex I might be leaving soon. 
 \label{ex:textbookgr1}
\end{exe}

\begin{exe}
 \ex I saw John on Sunday.
 \label{ex:textbookgr2}
\end{exe}

\begin{exe}
 \ex I really hate you right now.
 \label{ex:journalgr1}
\end{exe}

\begin{exe}
 \ex The guard made the prisoner unhappy.
 \label{ex:journalgr2}
\end{exe}

\hspace{0.5cm}

\section*{Bibliographical References}\label{sec:reference}

\bibliographystyle{lrec-coling2024-natbib}
\bibliography{lrec-coling2024-example}

\end{document}